# Semi-Peaucellier Linkage and Differential Mechanism for Linear Pinching and Self-Adaptive Grasping *

Haokai Ding, Zhaohan Chen, Tao Yang and Wenzeng Zhang, *Member, IEEE*

*Abstract*—This paper presents the SP-Diff parallel gripper system, addressing the limited adaptability of conventional end-effectors in intelligent industrial automation. The proposed design employs an innovative differential linkage mechanism with a modular symmetric dual-finger configuration to achieve linear-parallel grasping. By integrating a planetary gear transmission, the system enables synchronized linear motion and independent finger pose adjustment while maintaining structural rigidity, reducing Z-axis recalibration requirements by 30% compared to arc-trajectory grippers. The compact palm architecture incorporates a kinematically optimized parallelogram linkage and Differential mechanism, demonstrating adaptive grasping capabilities for diverse industrial workpieces and deformable objects such as citrus fruits. Future-ready interfaces are embedded for potential force/vision sensor integration to facilitate multimodal data acquisition (e.g., trajectory planning and object deformation) in digital twin frameworks. Designed as a flexible manufacturing solution, SP-Diff advances robotic end-effector intelligence through its adaptive architecture, showing promising applications in collaborative robotics, logistics automation, and specialized operational scenarios.

## I. INTRODUCTION

Automation represents a crucial frontier in contemporary technological advancement. As a core component of industrial automation and intelligent manufacturing, robotic hands are profoundly transforming traditional manufacturing paradigms through technological innovation and scenario expansion. They demonstrate significant advantages in enhancing efficiency, ensuring safety, and enabling flexible production, while progressively evolving toward intelligent, modular, and collaborative capabilities.

Robotic hands fall into three primary categories: industrial grippers, dexterous anthropomorphic hands, and underactuated hands.

Industrial grippers, widely deployed in industrial automation and intelligent manufacturing, feature substantial gripping force but remain constrained to arc-based parallel-pinching operations. Dexterous anthropomorphic hands achieve high-fidelity replication of human hand morphology through multi-joint biomimicry, exemplified by the Stanford/JPL Hand [1], Utah/MIT Hand [2], DLR/HIT Hand [3], Robonaut Hand [4], and CBPCA Architecture [5].

Underactuated hands, utilizing fewer actuators to achieve multiple degrees of freedom (DOF), serve as a cost-effective alternative to dexterous hands.

Despite their exceptional anthropomorphic capabilities, dexterous hands suffer from control complexity, excessive actuator counts, and prohibitive costs, limiting their viability as replacements for industrial grippers. Conversely, conventional industrial grippers exhibit inherent limitations. This dichotomy has propelled underactuated hands—with their simplified actuation, reduced motorization, and streamlined control—to the forefront of industrial automation research. Pioneering efforts in this domain include the early development of the MARS Hand [6] by Gosselin's team, the SARAH Hand [7] deployed on the International Space Station, and Gifu University's Gifu Hand [8]. Subsequent advancements further expanded the field, such as the SDM Hand by Prof. Dollar's group [9], the Omega Hand integrating passive compliance and underactuation for asymmetric scooping [10], and the tendon-driven Shadow Hand [11] by Shadow Robot Company. Tsinghua University's Indirect and Coupled Adaptive Transmission series [12], [13], and Tesla's six-actuator tendon-driven hand [14] controlling 11 joints further exemplify the evolution of underactuated designs. Commercial innovations like Robotiq's differential linkage mechanism for adaptive parallel grasping underscore the growing industrial relevance of these systems.

This paper proposes a novel Semi-Peaucellier mechanism derived from optimized Peaucellier linkage design, subsequently implemented in an underactuated differential adaptive robotic hand (Fig. 1). Our design achieves hybrid-mode switching between linear parallel pinching and adaptive grasping through enhanced Peaucellier linkages and differential mechanisms. Section II details grasping mode implementation principles, Section III describes the SP-Diff hand mechanism design, Section IV presents kinematic and mechanical analyses, Section V reports grasping experiments, and Section VI concludes the work.

* Research supported by the *Foundation of Enhanced Student Research Training (E-SRT)* and *Open Research for Innovation Challenges (ORIC)*, X-Institute.

Haokai Ding is with Future Technology School, Shenzhen Technology University and Laboratory of Robotics, X-Institute, Shenzhen, China.

Zhaohan Chen is with Future Technology School, Shenzhen Technology University, Shenzhen, China.

Tao Yang is with Future Technology School, Shenzhen Technology University, Shenzhen, China. (Corresponding author, email: yangtao@sztu.edu.cn)

Wenzeng Zhang is with Laboratory of Robotics, X-Institute, Shenzhen, China and Dept. of Mechanical Engineering, Tsinghua University, Beijing, China. (Corresponding author, email: zhangwenzeng@x-institute.edu.cn).

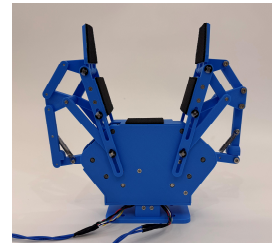

Figure 1. The SP-Diff gripper.

## II. DESIGN OF THE SP MECHANISM

### A. Linear Fingertip Motion: Semi-Peaucellier Linkage

The classical Peaucellier mechanism [15], depicted in Fig. 2a, employs an eight-bar linkage system to produce exact linear motion at point D through the propagation of geometric constraints. Kinematic analysis verifies that this mechanism achieves a straight-line trajectory owing to its unique linkage configuration.

The triangles $\Delta EC_1C_2$, $\Delta BC_1C_2$, and $\Delta DC_1C_2$ are all isosceles, sharing $C_1C_2$ as their common base, with vertices E, B, and D positioned as apexes. These apex points inherently lie on the perpendicular bisector of $C_1C_2$, ensuring their collinear alignment along this line due to the shared geometric constraint. This bisector functions simultaneously as the altitude, angle bisector, and median for each isosceles triangle, extending from the apex to the base. Similarly, for triangles $\Delta EFC_1$ and $\Delta DFC_1$, analogous geometric relationships can be established based on their vertex angles and congruent sides.

$$EC_1^2 = FE^2 + FC_1^2 \quad (1)$$

$$DC_1^2 = DF^2 + FC_1^2 \quad (2)$$

$$EC_1^2 - DC_1^2 = FE^2 - DF^2 = DE \times BE \quad (3)$$

Given the constant lengths of $EC_1$ and $DC_1$, the product $DE \times BE$ remains invariant—a critical geometric condition enabling precise straight-line motion. As a result, point D follows a strictly linear path orthogonal to AE throughout its motion.

The classical Peaucellier linkage faces inherent practical limitations due to its rigid geometric constraints: the enforced collinear alignment of nodes D-B-E (dictated by the trajectory of point E) introduces kinematic conflicts that compromise the precision of linear motion. To resolve this challenge, a Semi-Peaucellier linkage (SP) is proposed, achieved through targeted topological redesign as shown in Fig. 2b.

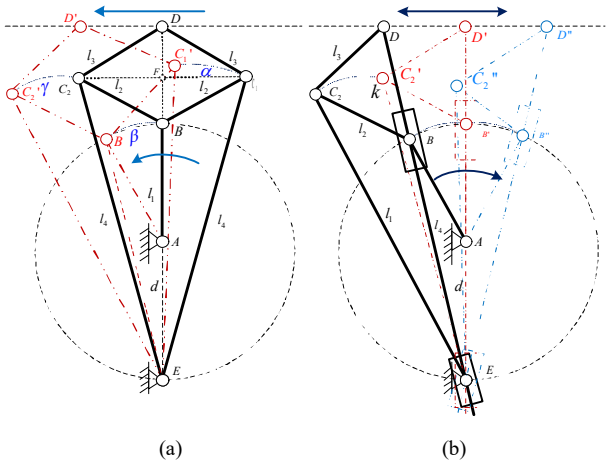

Figure 2. Peaucellier linkage and the Semi-Peaucellier mechanism.

This reengineered design consolidates the collinear nodes D-B-E into a single ternary link while incorporating sliding joints at B and E. Experimental studies demonstrated that gravitational effects induce vertical drift in the sliding joints, degrading linear motion precision. To counteract this displacement, a tension spring is integrated between links $l_7$ and $l_8$ to maintain positional stability of the sliders. The resulting SP configuration streamlines the traditional seven-bar design into a five-component system, enhancing structural efficiency without sacrificing motion accuracy.

The topological redesign achieves a 37.5% reduction in component count while resolving mechanical interference issues, enabling practical deployment of linear parallel clamping systems—a capability previously unattainable with conventional linkage architectures.

### B. Orientation-Stable Fingertip Control: Parallelogram Linkage Synthesis for Parallel Grasping

The Semi-Peaucellier Underactuated Grasping Finger mechanism was developed to maintain a fixed orientation of the gripper segment relative to the base during linear translation, enabling parallel grasping functionality. As shown in Fig. 3, the fingertip executes strictly linear motion across its operational range, ensuring the gripper's orientation remains invariant—a critical feature for stable and accurate object manipulation. This is achieved through a double-parallelogram mechanism, where two serially connected, geometrically identical parallelograms eliminate rotational motion while preserving pure translational displacement.

The optimized design simplifies control system requirements by minimizing angular displacement during grasping, thereby enhancing operational reliability. Its symmetrical kinematic architecture ensures balanced load distribution across gripper components, reducing susceptibility to misalignment under dynamic loads. Experimental evaluations confirm the mechanism's exceptional ability to maintain planar orientation integrity, making it particularly suitable for high-precision applications such as microscopic assembly systems and delicate material handling devices.

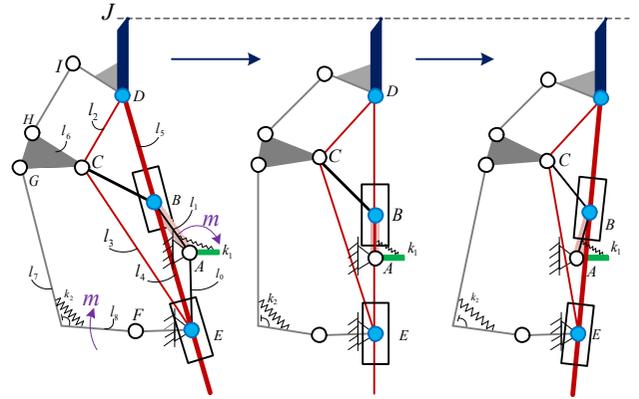

Figure 3. Structure and Linear Motion of the SP-Diff finger.

Furthermore, the mechanism's compact and lightweight architecture facilitates seamless integration into diverse robotic platforms without compromising overall payload or structural complexity. The theoretical framework and experimental validation of this system are elaborated in later sections of this work, offering a detailed exploration of its engineering rationale and functional benefits for real-world applications.

## III. DESIGN OF THE SP-DIFF GRIPPER

The SP-Diff gripper comprises two symmetrically configured grasping digits, with their actuation principles to be systematically examined in subsequent chapters. This symmetrical kinematic architecture ensures balanced load distribution and repeatable operational characteristics during object manipulation phases, a fundamental requirement for maintaining system stability and millimeter-level positioning precision. Furthermore, the kinematically synchronized digit configuration enables morphological compliance with polymorphic workpiece profiles while preserving high-fidelity contact retention throughout grasping cycles. The forthcoming analytical framework will delineate the governing principles of both trajectory planning algorithms and contact dynamics that define the operational envelope of the articulated mechanism.

### A. Structural Configuration

The finger mechanism integrates two core components: a semi-Peaucellier linkage and a reconfigured double parallelogram mechanism (DPM), as illustrated in Figure 4.

The mechanical system consists of three main parts: the proximal phalanx (1st P), the distal phalanx (2nd P), and the differential linkage mechanism. Through the distal joint axis, the 1st P and 2nd P are kinematically connected, enabling the 2nd P to achieve planar translation relative to the fixed base while maintaining autonomous rotational capability.

During static operation, the torsion spring S2 is placed within the driving linkage, preset at a 60° angle relative to the base, eliminating the need for planetary gears to achieve controlled circular motion. The use of the torsion spring significantly reduces gravitational interference on the linear guide rails Slat 1–Slat 2, thereby enabling stable linear motion generation in the SP actuator assembly.

The actuation of the SP-Diff module is achieved through the interaction between the spring in driving linkage 1 (DL1) and the limit block.

The coordinated interaction between the central joint axis and the proximal joint axis of ML1 ensures precise linear motion transmission, with both components constrained to slide within precision-machined guide channels. The operation of DL transmits force through the intermediate linkage ML4, generating programmable linear displacement in the 2nd P.

The compression spring S1, positioned between driving linkage 1 and driving linkage 2, restricts their relative motion during linear parallel pinching, maintaining them as a unified structure—ensuring the stability of the triangle formed by their three points. This stabilizes the reconfigured double parallelogram mechanism, effectively compensating for kinematic gaps in the bidirectional linear driving sequence while ensuring motion consistency. At the end of linear parallel pinching, the motor continues to drive the planetary gear, causing relative displacement between driving linkage 1 and driving linkage 2, enabling adaptive grasping.

### B. Differential mechanism

The differential parallel mechanism (DPM) integrates an optimized four-bar linkage system featuring extended structural members long link 1/ long link 2 paired with intermediate connectors $ML_2/ML_3$, where $ML_3$ adopts a triangular prism geometry. This constrained parallel kinematic configuration provides millimeter-level motion tracking accuracy through coordinated linkage movements.

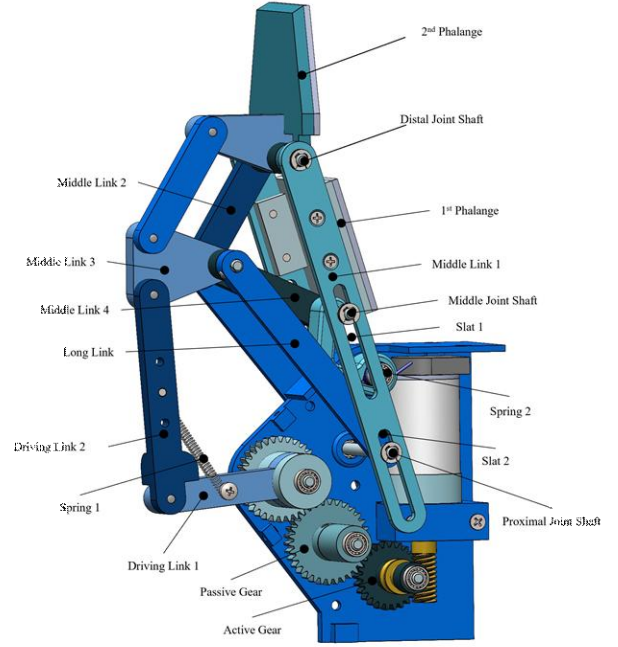

Figure 4. The Design of the SP-Diff finger.

The SP-Diff mechanism achieves adaptive enveloping of objects through active control, as illustrated in Figure 5(b). Computational analysis reveals that during the initial motion, the torsion spring pre-set in the driving linkage has an angle of 60°.

In standard parallel grasping operations, the motor gradually reduces this angular spacing until the finger fully closes, at which point the double-toggle blocks establish mechanical contact. This kinematic interaction drives DL1 to actuate DL2, causing deformation in the parallelogram linkage. Consequently, intermediate linkage 4 (ML4) undergoes compressive loading, transmitting the actuation force to the DL. This force transmission mechanism ultimately induces pronation of the entire finger assembly, enabling active adaptability.

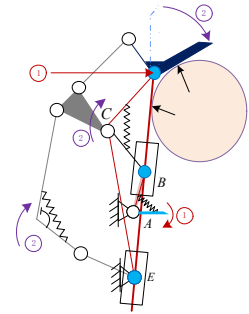

(a) Self-adaptive grasping

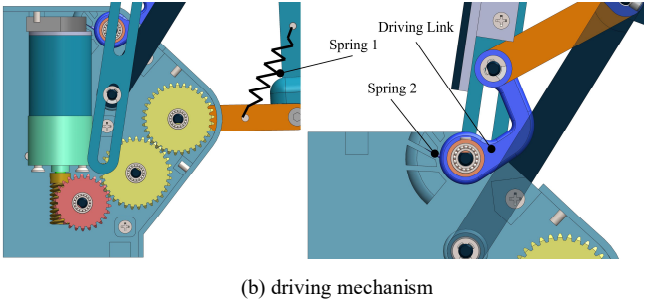

(b) driving mechanism

Figure 5. Self-adaptation and the driving mechanism, the differential mechanism between DL 1 and DL 2.

## C. Operational Modes

The SP-Diff gripper demonstrates dual-mode grasping capability, as illustrated in Fig. 6: a) The parallel grasping mode (Fig. 6a) is activated when manipulating relatively regular elongated objects and small standardized-shaped items, maintaining parallel fingertip alignment through constrained kinematic chains; b) The active-adaptive grasping mode (Fig. 6b) operates during handling of large irregular objects, exhibiting passive compliance through active self-adjusting rotation of secondary phalanges to achieve optimal shape conformity during experimental evaluations. These two operational modes ensure stable object interaction through distinct mechanical constraints and compliance mechanisms.

The finger mechanism demonstrates two distinct operational modes, as shown in Fig. 7. The underlying kinematic analysis reveals:

The SP-Diff finger mechanism incorporates dual transmission systems. The primary transmission system - comprising a worm gear set, active gear, passive gear, and actuation rod - implements parallel grasping functionality. The motor transmits torque through the worm gear set to the active gear, which subsequently engages the passive gear to drive the actuation rod. The arc motion of the actuation rod is converted into linear motion of the secondary phalange.

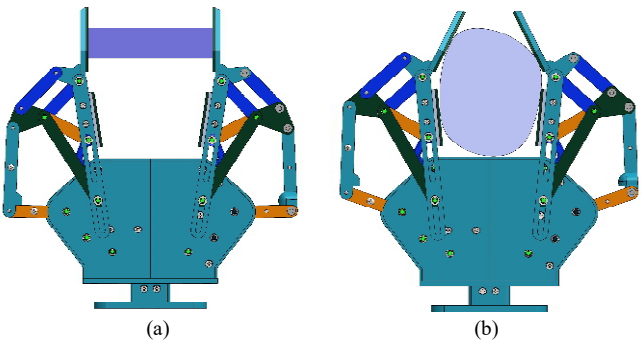

Figure 6. Two grasping modes of the SP-Diff gripper. (a) Linear parallel pinching for precision grasp; (b) Self-adaptive grasping for power grasp.

## IV. RESEARCH ON THE MAXIMUM GRASPING MASS OF PARALLEL JAW GRIPPERS

When a dual-fingered robotic hand grasps an arbitrarily oriented object in linear parallel pinching mode, its force model is illustrated in Fig. 7. To simplify the model, the grasped object is assumed to have uniform mass distribution and certain shape symmetry. In Fig. 7, $f$ represents the friction force between a single finger and the grasped object; $d$ denotes the distance from the finger-object contact point to the object's center of gravity; $G$ is the gravitational force on the object; $\alpha$ is the angle between the object's axis of symmetry and the vertical direction; $F$ is the normal force between the finger and object; and $\mu$ is the friction coefficient. $T$ is the frictional torque between the finger surface and object. $T \neq 0$ when the finger surface exhibits flexibility, whereas $T \approx 0$ when the finger surface has high rigidity with negligible torsional deformation. This section aims to establish a general mathematical model for linear parallel pinching, thus accounting for potential torque $T$ between the finger and object.

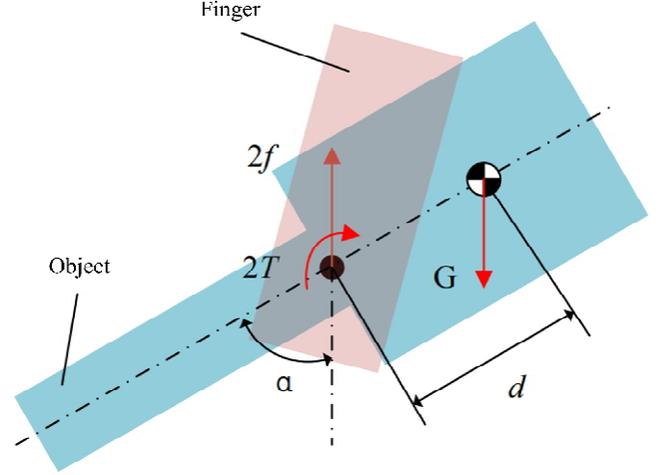

Figure 7. Linear Parallel-Jaw Simplified Model.

After mathematical derivation, it can be obtained that the frictional force generated by the fingertip should be:

$$f^2 + \frac{T^2}{e^2} \leq \mu^2 F_N^2 \quad (4)$$

where $e$ is the ratio of the maximum frictional torque to the maximum frictional force between a single finger and the object, and its expression is:

$$e = \frac{\max T}{\max f} \quad (5)$$

Through the above derivation, it can be concluded that when $d > 0$ and $2\mu F_n > G$, the conditions for the fingers to achieve stable grasping in the linear parallel-jaw mode are:

$$d \leq \max T \sqrt{\frac{4\mu^2 F_n^2 - G^2}{G^2 \sin^2 \alpha \mu^2 F_n^2}} \quad (6)$$

When $d = 0$, the frictional torque between the finger and the object is zero, and the condition for linear parallel-jaw grasping becomes:

$$2\mu F_n \geq G \quad (7)$$

When $d \neq 0$, the normal force $F_n$ that the finger needs to provide for stable linear parallel-jaw grasping is a function of $d$, $G$, $\mu$, and $\alpha$, varying with changes in these parameters. On the other hand, for a specific robotic hand, when the normal force that can be provided belongs to a certain specific range, according to the established model, the maximum grasping

mass for stable linear parallel-jaw grasping can be obtained as and shown in Fig. 8:

$$\max \text{mass} = \frac{1}{g}\sqrt{\frac{4\mu^2 F_n^2}{1+\frac{\mu^2 F_n^2 d^2 \sin^2 \alpha}{(\max T)^2}}} \quad (8)$$

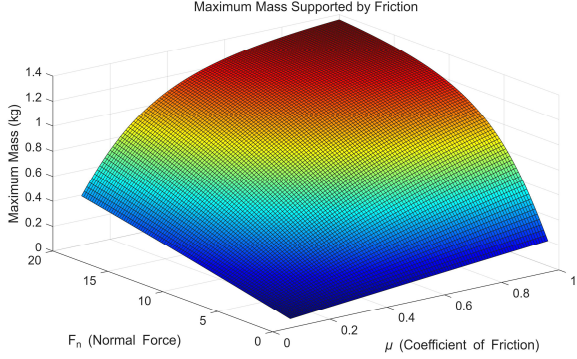

Figure 8. Maximum Mass Supported by Friction.

Table I lists the parameters used in the formula for calculating the maximum mass supported by friction. It includes the parameter name, description, range of values, and unit of measurement. This table helps in understanding the variables involved in the formula and their respective values used in the MATLAB simulation.

TABLE I. PARAMETERS USED IN MASS FORMULA

| Parameter | Description | Range | Unit |
|---|---|---|---|
| $\mu$ | Coefficient of friction | 0.1 to 1.0 | \ |
| $F_n$ | Normal force | 1 to 20 | N |
| d | Distance | 1 | M |
| $\alpha$ | Angle | $\pi/4$ | Radians |
| max T | Maximum tension | 5 | N |
| g | Acceleration due to gravity | 9.81 | m/s² |

In the adaptive enveloping grasping mode, the contact force of the two-segment underactuated robotic finger is expressed as:

$$f = J^{-T} P^{-T} t \quad (9)$$

where $f = [f_1 \; f_2]$ is the matrix expression of the contact forces corresponding to the two finger segments; $t = [T_a \; T_2]$ is the matrix expression of the torque generated by the actuator ($T_a$) and the torque generated by the passive element ($T_2$); $J$ is the Jacobian matrix of the finger grasping process, and $T$ is the transmission matrix determined by the finger mechanism used to transfer motion.

For a two-finger segment driven by a linkage mechanism, the corresponding Jacobian matrix and transmission matrix are:

$$J = \begin{bmatrix} k_1 & 0 \\ k_2 + l_1(\cos\theta_2 + \mu\sin\theta_2) & k_2 \end{bmatrix}, \quad T = \begin{bmatrix} 1 & R \\ 0 & 1 \end{bmatrix} \quad (10)$$

Where

$$R = -\frac{h}{h+l_1} \quad (11)$$

Combining the above equations, the matrix expression for the contact force can be obtained as:

$$f = \begin{bmatrix} \frac{l_1(k_2 - h(\cos\theta_2 + \mu\sin\theta_2))}{k_1 k_2 (h+l_1)} T_a - \frac{k_2 + l_1(\cos\theta_2 + \mu\sin\theta_2)}{k_1 k_2} T_2 \\ \frac{h}{k_2(h+l_1)} T_a + \frac{1}{k_2} T_2 \end{bmatrix} \quad (12)$$

Based on the above equations and the actual grasping conditions, the grasping forces of each finger segment can be calculated. Then, the stability of the finger's grasping state can be determined according to the conditions of force closure or form closure.

V. GRASPING EXPERIMENTS OF THE SP-DIFF GRIPPER

To systematically validate the design feasibility, a functional prototype was developed with rigorous consideration of kinematic compatibility and manufacturing tolerances. Key structural components—including the housing, base, and anthropomorphic fingers—were manufactured via fused deposition modeling (FDM) using a Bambu Lab A1 3D printer. Polylactic acid (PLA) material was selected for its optimal balance between cost-effectiveness and functional performance. A layer resolution of 0.1 mm and 40% hexagonal infill configuration achieved high dimensional accuracy.

Surface interaction performance was optimized by bonding 2-mm-thick silicone pads to the contact surfaces of the primary phalanx (proximal phalanx) and secondary phalanx (distal phalanx). Compared to bare PLA surfaces, this biomimetic treatment significantly enhanced grasping stability. The flexible interface maintains deformation compliance for adapting to irregular objects while simulating biomechanical properties of human fingertips.

Parallel pinching mode is realized through precise alignment of the selective parallel (SP) mechanism and double parallelogram mechanism, which maintains constant gripper orientation during linear displacement. This operational characteristic is critical for high-precision applications such as pick-and-place operations in industrial contexts and manipulation of delicate objects in medical robotics.

The SP-Diff gripper employs dual independent drive motors to actuate symmetrically arranged finger structures, providing dimensional adaptability and balanced force distribution. Its differential active-adaptive mechanism achieves stable shape-conforming envelopment. The design integrates a semi-Peaucellier linkage with a double parallelogram architecture, simultaneously fulfilling industrial automation requirements for precise linear motion and service robotics needs for adaptive grasping.

As shown in Fig. 9, the gripper successfully handles various objects (spherical, prismatic, cylindrical) through dual-mode coordination: The parallel grasping mode maintains fingertip linear alignment for symmetric targets, while the adaptive grasping mode achieves geometrically conforming stable retention of irregular shapes.

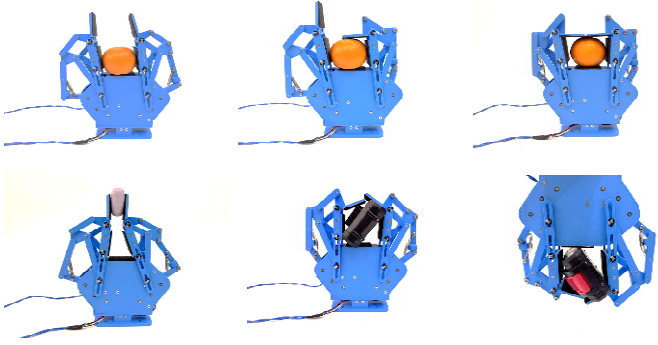

Figure 9.  Experiments of SP-Diff gripper.

The adaptive grasping of the SP-Diff gripper was experimentally validated using specified test objects. Results demonstrate the secondary phalange adaptively conforms to object profiles with simultaneous primary phalange tilting, confirming adaptive mechanics. This behavior stems from passive compliance through delayed active adaptation and precise linkage control, enabling geometric reconfiguration.

## VI.  Conclusions

This paper introduces a novel linear mechanism—the SP mechanism—which represents an optimized variation of the classic Peaucellier mechanism. Building upon this foundation, we present the SP-Diff gripper, a new underactuated adaptive robotic gripper equipped with two SP-Diff fingers. The SP-Diff finger integrates the parallel pinching mode commonly used in industrial grippers, achieving seamless integration of underactuation with single-motor control of two phalanges. Furthermore, the SP-Diff finger incorporates improvements to the conventional Peaucellier mechanism to enable linear parallel pinching. The design integrates differential linkage transmission, empowering the SP-Diff gripper to achieve adaptive enveloping grasping. Notably, the SP-Diff finger demonstrates impressive grasping force and exceptional adaptability to objects of diverse shapes.